\documentclass[conference]{IEEEtran}
\IEEEoverridecommandlockouts
\usepackage{placeins}
\usepackage{cite}
\usepackage{amsmath,amssymb,amsfonts}
\usepackage{algorithmic}
\usepackage{graphicx}
\usepackage{textcomp}
\usepackage{xcolor}
\definecolor{citeblue}{RGB}{87,145,220}
\usepackage[colorlinks=true,citecolor=citeblue]{hyperref}
\usepackage{booktabs}
\usepackage{multirow}
\usepackage{tabularx}
\def\BibTeX{{\rm B\kern-.05em{\sc i\kern-.025em b}\kern-.08em
    T\kern-.1667em\lower.7ex\hbox{E}\kern-.125emX}}
\usepackage{afterpage}
\usepackage{stfloats}
\begin{document}

\title{
PrismGS: Physically-Grounded Anti-Aliasing for High-Fidelity Large-Scale 3D Gaussian Splatting
}

\author{
Houqiang Zhong$^{1*}$,
Zhenglong Wu$^{2*}$,
Sihua Fu$^{2*}$,
Zihan Zheng$^{2}$, 
Xin Jin$^{3}$,
Xiaoyun Zhang$^{2}$,
Li Song$^{1\dagger}$,
Qiang Hu$^{2\dagger}$ \\
$^{1}$ School of Information Science and Electronic Engineering, Shanghai Jiao Tong University, Shanghai, China \\
$^{2}$ Cooperative Mediant Innovation Center, Shanghai Jiao Tong University, Shanghai, China\\
$^{3}$ College of Information Science and Technology, Eastern Institute of Technology, Ningbo, China \\
\{song\_li,qiang.hu\}@sjtu.edu.cn
\thanks{This work is supported by National Natural Science Foundation of China (62571322, 62431015, 62271308), STCSM(24ZR1432000, 24511106902, 24511106900, 22DZ2229005), 111 plan(BP0719010), and State Key Laboratory of UHD Video and Audio Production and Presentation. $^*$These authors contribute equally. $^\dagger$Corresponding authors.}
}

\maketitle

\begin{abstract}
3D Gaussian Splatting (3DGS) has recently enabled real-time photorealistic rendering in compact scenes, but scaling to large urban environments introduces severe aliasing artifacts and optimization instability, especially under high-resolution (e.g., 4K) rendering. These artifacts, manifesting as flickering textures and jagged edges, arise from the mismatch between Gaussian primitives and the multi-scale nature of urban geometry. While existing ``divide-and-conquer'' pipelines address scalability, they fail to resolve this fidelity gap. In this paper, we propose PrismGS, a physically-grounded regularization framework that improves the intrinsic rendering behavior of 3D Gaussians. PrismGS integrates two synergistic regularizers. The first is pyramidal multi-scale supervision, which enforces consistency by supervising the rendering against a pre-filtered image pyramid. This compels the model to learn an inherently anti-aliased representation that remains coherent across different viewing scales, directly mitigating flickering textures. This is complemented by an explicit size regularization that imposes a physically-grounded lower bound on the dimensions of the 3D Gaussians. This prevents the formation of degenerate, view-dependent primitives, leading to more stable and plausible geometric surfaces and reducing jagged edges. Our method is plug-and-play and compatible with existing pipelines. Extensive experiments on MatrixCity, Mill-19, and UrbanScene3D demonstrate that PrismGS achieves state-of-the-art performance, yielding significant PSNR gains around 1.5 dB against CityGaussian, while maintaining its superior quality and robustness under demanding 4K rendering.
\end{abstract}


\section{Introduction}

Recent advances in 3D Gaussian Splatting (3DGS)~\cite{3dgs} have redefined the frontier of radiance field~\cite{nerf,jointrf,4DGC,varfvv,VRVVC,hpc}, offering real-time photorealistic rendering for compact, object-centric scenes. However, extending this capability to large-scale, unbounded urban environments presents two fundamental challenges: scalability and aliasing. Urban scenes exhibit massive geometric complexity and long-range visibility, which exacerbate rendering artifacts such as flickering textures and jagged edges, especially under high-resolution (e.g., 4K) rendering. These aliasing issues not only degrade perceptual quality but also limit the practical deployment of 3DGS in applications like digital twins, autonomous simulation, and XR-based city modeling.

The evolution toward large-scale reconstruction begins with NeRF-based methods~\cite{nerf,blocknerf,gpnerf,UrbanRadianceFields,bungeenerf} like Mega-NeRF~\cite{MegaNERF}, which pioneer modular divide-and-conquer paradigms to address the spatial complexity of city blocks.
However, their reliance on implicit volumetric representations results in slow training and rendering speeds. This limitation prompts a community-wide shift toward explicit representations, notably 3DGS~\cite{3dgs} and its variant 2DGS~\cite{2dgs}.
Subsequent work focuses on adapting these explicit representations to city-scale scenes, primarily addressing the challenge of scalability. Octree-GS~\cite{octreegs} and others~\cite{scaffoldgs,Letsgo,contextgs,multiscalegs} employ hierarchical data structures to manage the large number of primitives. CityGaussian~\cite{citygaussian} and related approaches~\cite{vastgaussian,dogs,citygaussianv2}, adopt distributed block-wise optimization with Level-of-Detail strategies to improve efficiency. Momentum-GS~\cite{momentumgs} introduces a momentum-based self-distillation mechanism to improve consistency across independently trained blocks. Efficiency-oriented methods such as FlashGS~\cite{FlashGS,cityonweb,onthefly,scaleup3dgstraining,EfficientGS} significantly reduce training and rendering time.  However, aliasing artifacts, particularly under multi-scale viewing, remain pervasive and unresolved. In many high-resolution scenes, fine details shimmer, contours break, and surface coherence collapses, revealing a persistent gap in fidelity.

To bridge this fidelity gap, we introduce \textbf{PrismGS}, a regularization framework to mitigate aliasing in large-scale 3D Gaussian Splatting. Unlike previous efforts that focus on system-level scalability or architectural redesign, PrismGS directly enhances the intrinsic behavior of Gaussian primitives by enforcing consistency across scales and promoting geometric stability. Our key insight is that aliasing in urban-scale reconstructions arises from a mismatch between Gaussian parameters and the multi-scale nature of scene geometry and textures. 

\begin{figure}[h]
    \centering
    \includegraphics[width=\linewidth]{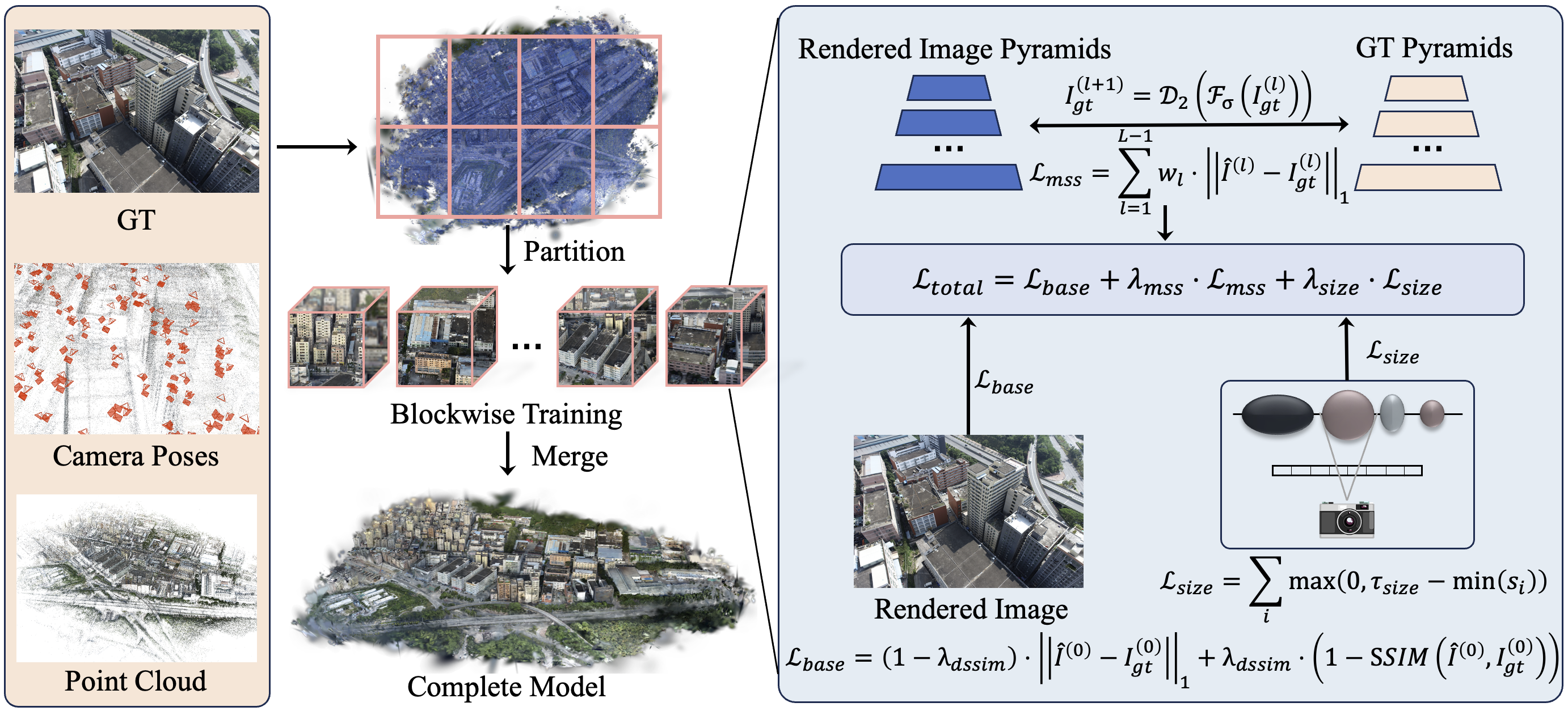}
    \caption{Overview of the PrismGS framework. Our method first partitions the scene into blocks for parallel training. During optimization, we introduce two key regularizers: a \textbf{Multi-Scale Supervision} loss for anti-aliasing and a \textbf{Size Regularization} loss for geometric stability.}
    \label{fig:overview}
\end{figure}

To address this issue, we jointly supervise rendering outputs across a range of resolutions during training, ensuring that each primitive contributes consistently at both coarse and fine scales. Simultaneously, we impose a physically-motivated constraint on the spatial extent of each Gaussian to prevent degenerate shapes and unstable optimization. These principles are integrated into a unified framework, regularizing the reconstruction process through both scale-consistent supervision and size-aware geometry control, effectively reducing artifacts such as flickering and jagged edges in high-resolution urban scenes. Extensive experiments demonstrate that PrismGS achieves state-of-the-art results in both quantitative metrics and perceptual quality, especially under 4K rendering conditions. Our contributions are summarized as follows:
\begin{itemize}
    \item We propose PrismGS, a physically-grounded regularization framework for large-scale 3DGS that improves anti-aliasing and rendering fidelity without compromising scalability.
    \item We introduce pyramidal multi-scale supervision for cross-resolution consistency, and Gaussian size regularization to enhance geometric stability and suppress high-frequency artifacts.
    \item Our method consistently outperforms existing approaches on challenging benchmarks, achieving \textbf{+1.0–1.5} dB PSNR gains and superior perceptual quality (SSIM, LPIPS) under high-resolution rendering.
\end{itemize}

\section{Methods}

We build upon 3DGS~\cite{3dgs}, which represents a scene as anisotropic Gaussians $G_i$ with position $\mu_i$, scale $\mathbf{s}_i$, rotation $\mathbf{q}_i$ (defining $\Sigma_i$), opacity $\alpha_i$, and SH color $\mathbf{c}_i$. Primitives are initialized from SfM~\cite{colmap} and rendered via a differentiable rasterizer with alpha blending. While efficient, direct application to city-scale scenes leads to cross-scale inconsistency and geometric degeneracy~\cite{citygaussian,momentumgs}. PrismGS addresses these two failure modes with multi-scale supervision and size regularization.

\begin{table*}[b]
    \centering 
    \caption{quantitative comparison across four large-scale scenes. We present metrics for PSNR$\uparrow$, SSIM$\uparrow$, and LPIPS$\downarrow$ on test views.}
    \label{tab:quant_comparison}
    \resizebox{0.95\linewidth}{!}{%
    \begin{tabular}{l|ccc|ccc|ccc|ccc}
        \toprule
        Scene & \multicolumn{3}{c|}{Building} & \multicolumn{3}{c|}{Rubble} & \multicolumn{3}{c|}{Residence} & \multicolumn{3}{c}{Sci-Art} \\
        \midrule
        Metrics & PSNR $\uparrow$ & SSIM $\uparrow$ & LPIPS $\downarrow$ & PSNR $\uparrow$ & SSIM $\uparrow$ & LPIPS $\downarrow$ & PSNR $\uparrow$ & SSIM $\uparrow$ & LPIPS $\downarrow$ & PSNR $\uparrow$ & SSIM $\uparrow$ & LPIPS $\downarrow$ \\
        \midrule
        Mega-NeRF  & 21.134 & 0.557 & 0.494 & 24.150 & 0.558 & 0.509 & \underline{22.054} & 0.632 & 0.485 & \textbf{25.719} & 0.772 & 0.386 \\
        2DGS  & 19.186 & 0.647 & 0.401 & 24.293 & 0.734 & 0.336 & 21.077 & 0.763 & 0.276 & 20.046 & 0.792 & 0.290 \\
        3DGS  & 20.437 & 0.716 & 0.304 & 25.304 & 0.787 & 0.262 & 21.697 & 0.789 & 0.228 & 21.644 & \underline{0.840} & 0.226 \\
        Octree-GS & 17.748 & 0.439 & 0.613 & 21.521 & 0.478 & 0.629 & 18.721 & 0.526 & 0.519 & 18.056 & 0.598 & 0.521 \\
        CityGaussian  & 21.483 & 0.757 & 0.268 & 24.929 & 0.772 & 0.268 & 21.720 & \underline{0.799} & 0.221 & 21.044 & 0.826 & 0.241 \\
        Momentum-GS & \underline{23.193} & \underline{0.810} & \underline{0.199} & \underline{25.771} & \underline{0.807} & \underline{0.227} & 22.040 & 0.798 & \underline{0.213} & 22.888 & 0.839 & \underline{0.222} \\
        \midrule
        Ours & \textbf{23.516} & \textbf{0.826} & \textbf{0.185} & \textbf{26.124} & \textbf{0.838} & \textbf{0.195} & \textbf{22.339} & \textbf{0.819} & \textbf{0.207} & \underline{23.317} & \textbf{0.851} & \textbf{0.209} \\
        \bottomrule
    \end{tabular}
    }
\end{table*}

\subsection{Tackling Aliasing with Multi-Scale Image Pyramids}
A primary challenge in large-scale rendering is aliasing, which manifests as flickering details and jagged edges when the scene is viewed from varying distances~\cite{citygaussian, 3dgs}. This occurs because rendering high-frequency geometry and textures at a coarse resolution without proper pre-filtering is analogous to undersampling a signal. An ideal coarse-level rendering, $\hat{I}_{\text{coarse}}$, should approximate a low-pass filtered version of the fine-level rendering, $\hat{I}_{\text{fine}}$~\cite{mipmap}. To enforce this constraint directly during optimization, we introduce a Multi-Scale Supervision (MSS) loss.

The core idea of MSS is to compel the model to maintain photometric consistency across a resolution pyramid, inspired by the classic mipmapping technique~\cite{mipmap}. During each training iteration, for a rendered image and its corresponding ground-truth image $I_{gt}$, we construct a pair of $L$-level image pyramids, $\mathcal{P}_{\text{render}} = \{\hat{I}^{(0)}, \dots, \hat{I}^{(L-1)}\}$ and $\mathcal{P}_{gt} = \{I_{gt}^{(0)}, \dots, I_{gt}^{(L-1)}\}$. To create a properly anti-aliased ground-truth pyramid, each level is generated by applying a low-pass filter (a Gaussian blur, denoted by the operator $\mathcal{F}_\sigma$) before downsampling (denoted by the operator $\mathcal{D}_s$ with a factor $s=2$). This process is defined as:
\begin{equation}
    I_{gt}^{(l+1)} = \mathcal{D}_2(\mathcal{F}_\sigma(I_{gt}^{(l)}))
\end{equation}
Here, the superscript $(l)$ denotes the pyramid level, with $l=0$ being the original resolution. The rendered pyramid $\{\hat{I}^{(l)}\}$ is produced by rendering the scene at each corresponding resolution. The MSS loss, $\mathcal{L}_{mss}$, is then formulated as the weighted sum of L1 norms across the downsampled levels:
\begin{align}
    \mathcal{L}_{mss} = \sum_{l=1}^{L-1} ||\hat{I}^{(l)} - I_{gt}^{(l)}||_1 .
\end{align}
By penalizing discrepancies at lower resolutions against a properly pre-filtered ground truth, this loss function acts as an implicit, end-to-end differentiable anti-aliasing filter. It forces the optimizer to learn a set of Gaussian parameters that are not only accurate for the high-resolution view but also remain stable and coherent when downsampled, effectively baking anti-aliasing properties into the 3D primitives themselves.

\subsection{Preventing Geometric Degeneracy with Size Regularization}
Another key challenge in 3DGS is geometric instability, where the optimization process, driven solely by a photometric loss, may produce physically implausible, degenerate primitives. These often take the form of extremely thin ``needle-like'' or flat ``pancake-like'' Gaussians that overfit to high-frequency details in the training images. Such primitives are not robust and cause rendering artifacts like holes and flickering when viewed from novel angles or under high magnification.

To address this, we introduce an explicit 3D Gaussian size regularization loss, $\mathcal{L}_{size}$. The goal is to prevent the model from creating primitives smaller than a physical limit. For each camera in the training set with focal length $f$, the pixel sampling interval in 3D space at a depth $d$ is $T=d/f$. According to the Nyquist theorem, to reconstruct a signal without aliasing, the smallest resolvable 3D structure is approximately $2T$. We can therefore establish a global minimum sampling interval, $T_{min}$, across all training views to define a physical lower bound on Gaussian size. The regularization loss penalizes any Gaussian whose smallest scaling axis falls below a defined threshold, $\tau_{size}$:
\begin{align}
    \mathcal{L}_{size} = \sum_{i} \max(0, \tau_{size} - \min(\mathbf{s}_{i}))
\end{align}
where $\tau_{size}$ is a hyperparameter defining the minimum allowable scaling axis, and $\min(\mathbf{s}_{i})$ is the smallest component of the scaling vector $\mathbf{s}_i$ for the $i$-th Gaussian. This loss effectively suppresses high-frequency artifacts and encourages the formation of smoother, more continuous surfaces that better represent the true scene geometry.

\subsection{Joint Optimization for Robust Reconstruction}
Our final training objective integrates the standard photometric loss with our two novel regularization terms. The base reconstruction loss, $\mathcal{L}_{base}$, is a weighted combination of an L1 norm and a structural dissimilarity (D-SSIM) loss, calculated at the highest resolution ($l=0$):
\begin{equation}
\resizebox{0.9\linewidth}{!}{$
    \mathcal{L}_{base} = (1-\lambda_{dssim}) \cdot ||\hat{I}^{(0)} - I_{gt}^{(0)}||_1
    + \lambda_{dssim} \cdot (1 - \text{SSIM}(\hat{I}^{(0)}, I_{gt}^{(0)}))
$}
\end{equation}

The total loss function, $\mathcal{L}_{total}$, is then a weighted sum of the base loss and our two regularization terms:
\begin{align}
    \mathcal{L}_{total} = \mathcal{L}_{base} + \lambda_{mss} \cdot \mathcal{L}_{mss} + \lambda_{size} \cdot \mathcal{L}_{size}
\end{align}
The hyperparameters $\lambda_{dssim}$, $\lambda_{mss}$, and $\lambda_{size}$ balance the influence of the structural dissimilarity, the multi-scale supervision, and the size regularization, respectively. This unified objective function guides the optimization to produce a 3D representation that is robust against both aliasing and geometric degradation, making it highly suitable for high-fidelity, large-scale scene reconstruction.

\section{Experiments}
Our framework builds on Momentum-GS~\cite{momentumgs} and is trained for 60,000 iterations on 8 NVIDIA 3090 GPUs. We fix all loss weights across experiments: $\lambda_{dssim}=0.2$, $\lambda_{mss}=0.1$, and $\lambda_{size}=0.01$. Quantitative and qualitative evaluations are conducted on three large-scale benchmarks: MatrixCity~\cite{matrixcity}, Mill-19~\cite{MegaNERF}, and UrbanScene3D~\cite{UrbanScene3D}. Following prior work~\cite{citygaussian}, all input images are downsampled by a factor of 4 for standard evaluation. To assess anti-aliasing performance under challenging conditions, we additionally render high-resolution ($3840 \times 2160$) novel views on the \texttt{Building} and \texttt{Rubble} scenes from Mill-19. We compare our method against SOTA approaches, including the NeRF-based Mega-NeRF~\cite{MegaNERF}, and Gaussian-based methods: 3DGS~\cite{3dgs}, 2DGS~\cite{2dgs}, Octree-GS~\cite{octreegs}, CityGaussian~\cite{citygaussian}, and Momentum-GS~\cite{momentumgs}. All these methods are evaluated using their default training strategies and hyperparameter configurations to ensure a fair comparison.

\subsection{Quantitative Comparisons}
As shown in Tab.~\ref{tab:quant_comparison}, PrismGS consistently outperforms prior methods. On the \texttt{Building} scene, our LPIPS of 0.185 marks a significant improvement over the baseline Momentum-GS's 0.199. This directly reflects the success of our anti-aliasing objective, as LPIPS is highly sensitive to the flickering and texture shimmering. This advantage is further magnified in our 4K high-resolution in Tab.~\ref{tab:4k_quantitative}. Here, PrismGS maintains its performance lead, outperforming all competitors across all metrics on the \texttt{Rubble} dataset. This robust performance at a demanding resolution directly validates the effectiveness of our physically-grounded regularization.

\begin{table}
\centering
\caption{quantitative comparison under 4k resolution.}
\label{tab:4k_quantitative}
\begin{tabular}{l|c|ccc}
\toprule
Dataset & Method &  PSNR$\uparrow$ & SSIM$\uparrow$ & LPIPS$\downarrow$ \\
\midrule
\multirow{6}{*}{Building Dataset} & 
Mega-NeRF & 20.078 & 0.537 & 0.619 \\
&3DGS & 17.937 & 0.589 & 0.481 \\
&Octree-GS & 17.344 & 0.513 & 0.603 \\
&CityGaussian & 20.523 & \underline{0.683} & \textbf{0.398} \\
&Momentum-GS & \underline{21.057} & 0.682 & 0.426 \\
 & ours & \textbf{21.291} & \textbf{0.693} & \underline{0.401} \\
\midrule
\multirow{6}{*}{Rubble Dataset} & 
Mega-NeRF & 22.873 & 0.516 & 0.656 \\
&3DGS & 23.457 & 0.643 & 0.483 \\
&Octree-GS & 20.887 & 0.516 & 0.625 \\
&CityGaussian & 23.436 & 0.661 & 0.452 \\
&Momentum-GS & \underline{23.717} & \underline{0.671} & \underline{0.422} \\
 & ours & \textbf{24.001} & \textbf{0.687} & \textbf{0.401} \\
\bottomrule
\end{tabular}
\end{table}

\subsection{Qualitative Comparisons}
\begin{figure*}[tb]
    \centering
    \includegraphics[width=\linewidth]{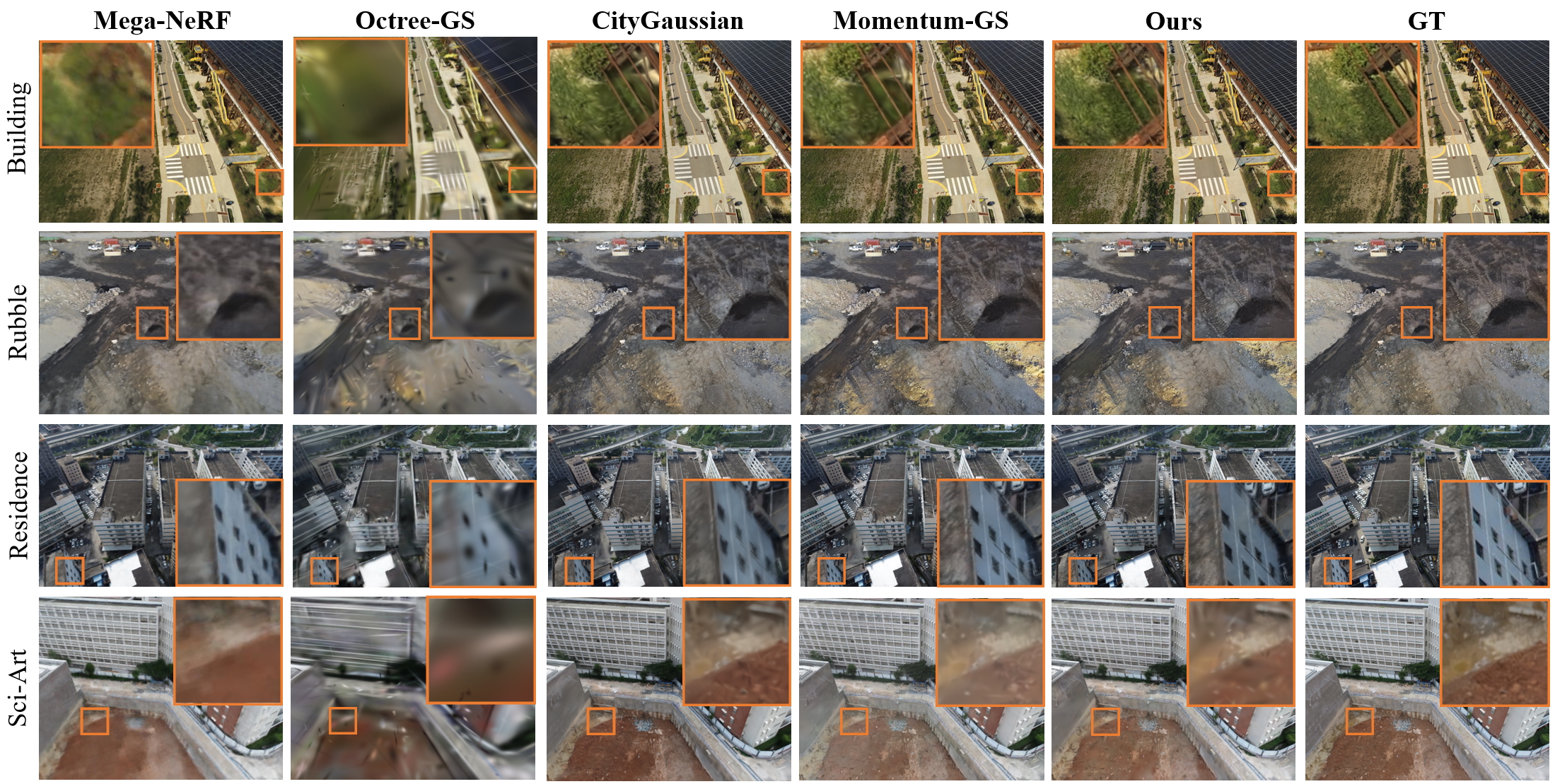}
    \vspace{-6mm}
    \caption{Qualitative comparisons of different methods (Mega-NeRF, Octree-GS, CityGaussian, Momentum-GS, Ours) against Ground Truth  across four large-scale scenes. Orange insets highlight patches that reveal notable visual differences, demonstrating the superiority of our method in capturing fine details and maintaining structural consistency.}
    \label{fig:scene_reconstruction}
\end{figure*}

The qualitative results in Fig.~\ref{fig:scene_reconstruction} further corroborate our quantitative findings. Visual comparison reveals that PrismGS generates renderings with substantially higher clarity and detail compared to other methods like CityGaussian and Momentum-GS. As highlighted in the magnified insets, our approach excels at reconstructing fine-grained textures and sharp geometric details on distant structures, whereas other methods often suffer from blurriness or aliasing artifacts. Furthermore, our method effectively preserves structural consistency and mitigates the visual popping artifacts common in large-scale rendering, confirming the benefits of our proposed pyramidal supervision and geometric regularization.

\subsection{Ablation Study}
\begin{figure}[htbp]
    \centering
    \includegraphics[width=0.95\columnwidth]{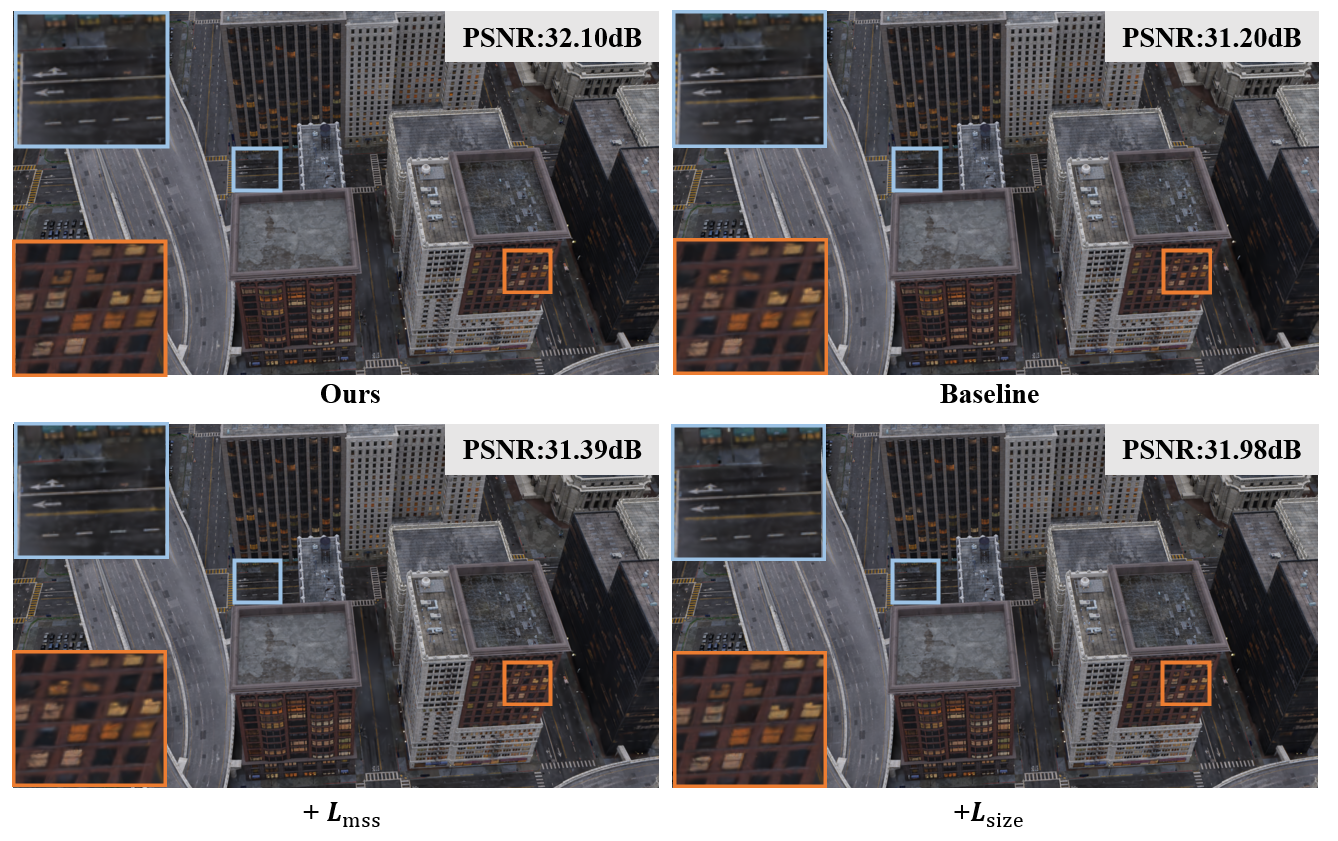}
    \vspace{-3mm}
    \caption{Qualitative results of ablation study. Excluding any module leads to lower reconstruction quality and other impacts.}
    \label{fig:ablation_qualitative}
\end{figure}

\begin{table}[ht]
\vspace{-4.5mm}
\centering
\caption{ablation study on different strategy of measuring the reconstruction quality under MatrixCity scene}
\label{tab:ablation_matrixcity}
\vspace{-3mm}
\begin{tabular}{c|ccc}
\toprule
Models & PSNR & SSIM & LPIPS \\
\midrule
baseline & 27.768 & 0.867 & 0.212 \\
$+ \mathcal{L}_{mss}$ & 27.892 & 0.870 & 0.206 \\
$+ \mathcal{L}_{size}$ & 28.124 & 0.878 & 0.182 \\
\midrule
Full model & \textbf{28.272} & \textbf{0.888} & \textbf{0.173} \\
\bottomrule
\end{tabular}
\end{table}

In Tab.~\ref{tab:ablation_matrixcity}, we conduct an ablation study on the MatrixCity scene.
Adding only $\mathcal{L}_{mss}$ primarily reduces LPIPS, indicating fewer aliasing artifacts. 
Adding only the size regularization ($\mathcal{L}_{size}$) yields a more substantial performance leap, boosting PSNR to 28.124 and significantly lowering LPIPS to 0.182. This highlights the critical role of constraining Gaussian sizes in preventing geometric degeneracy. 
Finally, our full model, which integrates both components, achieves the best performance, pushing PSNR to 28.272 and LPIPS to 0.173. 
This analysis confirms that our two modules are complementary: $\mathcal{L}_{mss}$ primarily targets view-dependent aliasing, while $\mathcal{L}_{size}$ enforces view-independent geometric stability.
The qualitative results of the ablation in Fig.~\ref{fig:ablation_qualitative} further support this conclusion: excluding any module will lead to a decrease in reconstruction quality and other impacts. Only a complete model can achieve the best performance.

\vspace{-2mm}
\section{Conclusion} 
In this paper, we introduced PrismGS, a regularization framework designed to significantly enhance the fidelity of large-scale 3D Gaussian Splatting. By building upon scalable block-based pipelines, our work specifically targets the pervasive issues of aliasing and geometric instability. PrismGS integrates two synergistic components: a pyramidal multi-scale supervision loss that enforces rendering consistency across different resolutions, and a physically-grounded size regularization that prevents the formation of degenerate, view-dependent primitives. Extensive experiments on challenging benchmarks demonstrate that our method achieves SOTA results, significantly improving both quantitative metrics and perceptual quality, especially for high-fidelity 4K rendering. A limitation of our method is its assumption of a static environment, as it does not explicitly filter dynamic objects. One promising direction is the integration of semantic scene understanding to differentiate and model static and dynamic elements separately. 

\clearpage
\bibliographystyle{IEEEtran}
\bibliography{reference}

\begin{thebibliography}{10}
\providecommand{\url}[1]{#1}
\csname url@samestyle\endcsname
\providecommand{\newblock}{\relax}
\providecommand{\bibinfo}[2]{#2}
\providecommand{\BIBentrySTDinterwordspacing}{\spaceskip=0pt\relax}
\providecommand{\BIBentryALTinterwordstretchfactor}{4}
\providecommand{\BIBentryALTinterwordspacing}{\spaceskip=\fontdimen2\font plus
\BIBentryALTinterwordstretchfactor\fontdimen3\font minus \fontdimen4\font\relax}
\providecommand{\BIBforeignlanguage}[2]{{%
\expandafter\ifx\csname l@#1\endcsname\relax
\typeout{** WARNING: IEEEtran.bst: No hyphenation pattern has been}%
\typeout{** loaded for the language `#1'. Using the pattern for}%
\typeout{** the default language instead.}%
\else
\language=\csname l@#1\endcsname
\fi
#2}}
\providecommand{\BIBdecl}{\relax}
\BIBdecl

\bibitem{3dgs}
B.~Kerbl, G.~Kopanas, T.~Leimkuehler, and G.~Drettakis, ``3d gaussian splatting for real-time radiance field rendering,'' \emph{ACM Trans. Graph.}, vol.~42, no.~4, July 2023.

\bibitem{nerf}
B.~Mildenhall, P.~P. Srinivasan, M.~Tancik, J.~T. Barron, R.~Ramamoorthi, and R.~Ng, ``Nerf: representing scenes as neural radiance fields for view synthesis,'' \emph{Commun. ACM}, vol.~65, no.~1, p. 99–106, December 2021.

\bibitem{jointrf}
Z.~Zheng, H.~Zhong, Q.~Hu, X.~Zhang, L.~Song, Y.~Zhang, and Y.~Wang, ``Jointrf: End-to-end joint optimization for dynamic neural radiance field representation and compression,'' in \emph{2024 IEEE International Conference on Image Processing (ICIP)}, 2024, pp. 3292--3298.

\bibitem{4DGC}
Q.~Hu, Z.~Zheng, H.~Zhong, S.~Fu, L.~Song, X.~Zhang, G.~Zhai, and Y.~Wang, ``4dgc: Rate-aware 4d gaussian compression for efficient streamable free-viewpoint video,'' in \emph{Proceedings of the Computer Vision and Pattern Recognition Conference (CVPR)}, June 2025, pp. 875--885.

\bibitem{varfvv}
Q.~Hu, Q.~He, H.~Zhong, G.~Lu, X.~Zhang, G.~Zhai, and Y.~Wang, ``Varfvv: View-adaptive real-time interactive free-view video streaming with edge computing,'' \emph{IEEE Journal on Selected Areas in Communications}, vol.~43, no.~7, pp. 2620--2634, 2025.

\bibitem{VRVVC}
Q.~Hu, H.~Zhong, Z.~Zheng, X.~Zhang, Z.~Cheng, L.~Song, G.~Zhai, and Y.~Wang, ``Vrvvc: Variable-rate nerf-based volumetric video compression,'' in \emph{Proceedings of the AAAI Conference on Artificial Intelligence}, vol.~39, no.~4, 2025, pp. 3563--3571.

\bibitem{hpc}
Z.~Zheng, H.~Zhong, Q.~Hu, X.~Zhang, L.~Song, Y.~Zhang, and Y.~Wang, ``Hpc: Hierarchical progressive coding framework for volumetric video,'' in \emph{Proceedings of the 32nd ACM International Conference on Multimedia}, 2024, pp. 7937--7946.

\bibitem{blocknerf}
M.~Tancik, V.~Casser, X.~Yan, S.~Pradhan, B.~P. Mildenhall, P.~Srinivasan, J.~T. Barron, and H.~Kretzschmar, ``Block-nerf: Scalable large scene neural view synthesis,'' in \emph{2022 IEEE/CVF Conference on Computer Vision and Pattern Recognition (CVPR)}, 2022, pp. 8238--8248.

\bibitem{gpnerf}
Y.~Zhang, G.~Chen, and S.~Cui, ``Efficient large-scale scene representation with a hybrid of high-resolution grid and plane features,'' \emph{Pattern Recognition}, vol. 158, p. 111001, 2025.

\bibitem{UrbanRadianceFields}
K.~Rematas, A.~Liu, P.~Srinivasan, J.~Barron, A.~Tagliasacchi, T.~Funkhouser, and V.~Ferrari, ``Urban radiance fields,'' in \emph{2022 IEEE/CVF Conference on Computer Vision and Pattern Recognition (CVPR)}, 2022, pp. 12\,922--12\,932.

\bibitem{bungeenerf}
Y.~Xiangli, L.~Xu, X.~Pan, N.~Zhao, A.~Rao, C.~Theobalt, B.~Dai, and D.~Lin, ``Bungeenerf: Progressive neural radiance field for extreme multi-scale scene rendering,'' in \emph{Computer Vision – ECCV 2022: 17th European Conference, Tel Aviv, Israel, October 23–27, 2022, Proceedings, Part XXXII}.\hskip 1em plus 0.5em minus 0.4em\relax Berlin, Heidelberg: Springer-Verlag, 2022, p. 106–122.

\bibitem{MegaNERF}
H.~Turki, D.~Ramanan, and M.~Satyanarayanan, ``Mega-nerf: Scalable construction of large-scale nerfs for virtual fly-throughs,'' in \emph{Proceedings of the IEEE/CVF Conference on Computer Vision and Pattern Recognition (CVPR)}, June 2022, pp. 12\,922--12\,931.

\bibitem{2dgs}
B.~Huang, Z.~Yu, A.~Chen, A.~Geiger, and S.~Gao, ``2d gaussian splatting for geometrically accurate radiance fields,'' in \emph{ACM SIGGRAPH 2024 Conference Papers}, ser. SIGGRAPH '24.\hskip 1em plus 0.5em minus 0.4em\relax New York, NY, USA: Association for Computing Machinery, 2024.

\bibitem{octreegs}
K.~Ren, L.~Jiang, T.~Lu, M.~Yu, L.~Xu, Z.~Ni, and B.~Dai, ``Octree-gs: Towards consistent real-time rendering with lod-structured 3d gaussians,'' \emph{IEEE Transactions on Pattern Analysis and Machine Intelligence}, pp. 1--15, 2025.

\bibitem{scaffoldgs}
T.~Lu, M.~Yu, L.~Xu, Y.~Xiangli, L.~Wang, D.~Lin, and B.~Dai, ``Scaffold-gs: Structured 3d gaussians for view-adaptive rendering,'' in \emph{Proceedings of the IEEE/CVF Conference on Computer Vision and Pattern Recognition}, 2024, pp. 20\,654--20\,664.

\bibitem{Letsgo}
J.~Cui, J.~Cao, F.~Zhao, Z.~He, Y.~Chen, Y.~Zhong, L.~Xu, Y.~Shi, Y.~Zhang, and J.~Yu, ``Letsgo: Large-scale garage modeling and rendering via lidar-assisted gaussian primitives,'' \emph{ACM Transactions on Graphics (TOG)}, vol.~43, no.~6, pp. 1--18, 2024.

\bibitem{contextgs}
Y.~Wang, Z.~Li, L.~Guo, W.~Yang, A.~Kot, and B.~Wen, ``Context{GS} : Compact 3d gaussian splatting with anchor level context model,'' in \emph{The Thirty-eighth Annual Conference on Neural Information Processing Systems}, 2024.

\bibitem{multiscalegs}
Z.~Yan, W.~F. Low, Y.~Chen, and G.~H. Lee, ``Multi-scale 3d gaussian splatting for anti-aliased rendering,'' in \emph{2024 IEEE/CVF Conference on Computer Vision and Pattern Recognition (CVPR)}, 2024, pp. 20\,923--20\,931.

\bibitem{citygaussian}
Y.~Liu, C.~Luo, L.~Fan, N.~Wang, J.~Peng, and Z.~Zhang, ``Citygaussian: Real-time high-quality large-scale scene rendering with gaussians,'' in \emph{Computer Vision – ECCV 2024: 18th European Conference, Milan, Italy, September 29–October 4, 2024, Proceedings, Part XVI}.\hskip 1em plus 0.5em minus 0.4em\relax Berlin, Heidelberg: Springer-Verlag, 2024, p. 265–282.

\bibitem{vastgaussian}
J.~Lin, Z.~Li, X.~Tang, J.~Liu, S.~Liu, J.~Liu, Y.~Lu, X.~Wu, S.~Xu, Y.~Yan, and W.~Yang, ``Vastgaussian: Vast 3d gaussians for large scene reconstruction,'' in \emph{Proceedings of the IEEE/CVF Conference on Computer Vision and Pattern Recognition (CVPR)}, June 2024, pp. 5166--5175.

\bibitem{dogs}
Y.~Chen and G.~H. Lee, ``Dogs: Distributed-oriented gaussian splatting for large-scale 3d reconstruction via gaussian consensus,'' in \emph{The Thirty-eighth Annual Conference on Neural Information Processing Systems}, 2024.

\bibitem{citygaussianv2}
Y.~Liu, C.~Luo, Z.~Mao, J.~Peng, and Z.~Zhang, ``Citygaussianv2: Efficient and geometrically accurate reconstruction for large-scale scenes,'' in \emph{The Thirteenth International Conference on Learning Representations}, 2025.

\bibitem{momentumgs}
J.~Fan, W.~Li, Y.~Han, and Y.~Tang, ``Momentum-gs: Momentum gaussian self-distillation for high-quality large scene reconstruction,'' in \emph{Proceedings of the IEEE/CVF International Conference on Computer Vision (ICCV)}, October 2025.

\bibitem{FlashGS}
G.~Feng, S.~Chen, R.~Fu, Z.~Liao, Y.~Wang, T.~Liu, B.~Hu, L.~Xu, Z.~Pei, H.~Li, X.~Li, N.~Sun, X.~Zhang, and B.~Dai, ``Flashgs: Efficient 3d gaussian splatting for large-scale and high-resolution rendering,'' in \emph{Proceedings of the Computer Vision and Pattern Recognition Conference (CVPR)}, June 2025, pp. 26\,652--26\,662.

\bibitem{cityonweb}
K.~Song, X.~Zeng, C.~Ren, and J.~Zhang, ``City-on-web: Real-time neural rendering of large-scale scenes on the web,'' in \emph{Computer Vision -- ECCV 2024}, A.~Leonardis, E.~Ricci, S.~Roth, O.~Russakovsky, T.~Sattler, and G.~Varol, Eds.\hskip 1em plus 0.5em minus 0.4em\relax Cham: Springer Nature Switzerland, 2025, pp. 385--402.

\bibitem{onthefly}
A.~Meuleman, I.~Shah, A.~Lanvin, B.~Kerbl, and G.~Drettakis, ``On-the-fly reconstruction for large-scale novel view synthesis from unposed images,'' \emph{{ACM Transactions on Graphics}}, vol.~44, no.~4, Aug. 2025, nef/OPAL.

\bibitem{scaleup3dgstraining}
H.~Zhao, H.~Weng, D.~Lu, A.~Li, J.~Li, A.~Panda, and S.~Xie, ``On scaling up 3d gaussian splatting training,'' in \emph{Computer Vision -- ECCV 2024 Workshops}, A.~Del~Bue, C.~Canton, J.~Pont-Tuset, and T.~Tommasi, Eds.\hskip 1em plus 0.5em minus 0.4em\relax Cham: Springer Nature Switzerland, 2025, pp. 14--36.

\bibitem{EfficientGS}
W.~Liu, T.~Guan, B.~Zhu, L.~Xu, Z.~Song, D.~Li, Y.~Wang, and W.~Yang, ``Efficientgs: Streamlining gaussian splatting for large-scale high-resolution scene representation,'' \emph{IEEE MultiMedia}, vol.~32, no.~1, pp. 61--71, 2025.

\bibitem{colmap}
J.~L. Schönberger and J.-M. Frahm, ``Structure-from-motion revisited,'' in \emph{2016 IEEE Conference on Computer Vision and Pattern Recognition (CVPR)}, 2016, pp. 4104--4113.

\bibitem{mipmap}
J.~P. Ewins, M.~D. Waller, M.~White, and P.~F. Lister, ``Implementing an anisotropic texture filter,'' \emph{Computers \& Graphics}, vol.~24, no.~2, pp. 253--267, 2000.

\bibitem{matrixcity}
Y.~Li, L.~Jiang, L.~Xu, Y.~Xiangli, Z.~Wang, D.~Lin, and B.~Dai, ``Matrixcity: A large-scale city dataset for city-scale neural rendering and beyond,'' in \emph{Proceedings of the IEEE/CVF International Conference on Computer Vision}, 2023, pp. 3205--3215.

\bibitem{UrbanScene3D}
L.~Lin, Y.~Liu, Y.~Hu, X.~Yan, K.~Xie, and H.~Huang, ``Capturing, reconstructing, and simulating: the urbanscene3d dataset,'' in \emph{ECCV}, 2022.

\end{thebibliography}

\end{document}